\documentclass[runningheads]{llncs}
\usepackage{graphicx}
\usepackage{multirow}
\usepackage{array}
\usepackage{booktabs}
\usepackage{amsmath,amssymb,amsfonts}
\usepackage{algorithmic}
\usepackage{algorithm}
\usepackage{textcomp}
\usepackage{xcolor}
\usepackage{array}
\usepackage{subfigure}

\begin{document}
%
\title{Personalized Federated Learning on Heterogeneous and Long-Tailed Data via Expert Collaborative Learning}
\titlerunning{Federated Expert Collaboration}
\author{Paper ID 754}
\institute{}
%
%
\author{Fengling Lv\inst{1} \and
Xinyi Shang\inst{2} \and
Yang Zhou\inst{3}\and
Yiqun Zhang\inst{4} \and
Mengke Li\inst{5}\and
Yang Lu\inst{1,6}
}

%
\institute{ Key Laboratory of Multimedia Trusted Perception and Efficient Computing, Xiamen University, China\\
\email{lvfenlging@stu.xmu.edu.cn}
\and
University College London,  England\\
\email{xinyi.shang.23@ucl.ac.uk}\\
\and
A*STAR Institute of High Performance Computing\\
\email{zhou\_yang@ihpc.a-star.edu.sg}
\and
School of Computer Science and Technology, Guangdong University of Technology, Guangzhou, China\\
\email{yqzhang@gdut.edu.cn}
\and
Guangdong Laboratory of Artificial Intelligence and Digital Economy, Shenzhen, China\\
\email{limengke@gml.ac.cn}
\and
Fujian Key Laboratory of Sensing and Computing for Smart City, School of Informatics, Xiamen University, China\\
\email{luyang@xmu.edu.cn}}

\maketitle              
\begin{abstract}
Personalized Federated Learning (PFL) aims to acquire customized models for each client without disclosing raw data by leveraging the collective knowledge of distributed clients. However, the data collected in real-world scenarios is likely to follow a long-tailed distribution. For example, in the medical domain, it is more common for the number of general health notes to be much larger than those specifically related to certain diseases.
The presence of long-tailed data can significantly degrade the performance of PFL models. Additionally, due to the diverse environments in which each client operates, data heterogeneity is also a classic challenge in federated learning. In this paper, we explore the joint problem of global long-tailed distribution and data heterogeneity in PFL and propose a method called Expert Collaborative Learning (ECL) to tackle this problem. Specifically, each client has multiple experts, and each expert has a different training subset, which ensures that each class, especially the minority classes, receives sufficient training. Multiple experts collaborate synergistically to produce the final prediction output. Without special bells and whistles, the vanilla ECL outperforms other state-of-the-art PFL methods on several benchmark datasets under different degrees of data heterogeneity and long-tailed distribution.
\end{abstract}
\section{Introduction}
\label{sec:intro}
In recent years, deep learning technology has made incredible progress in various fields, such as computer vision and natural language processing \cite{zhao2023federated}.
A common way of constructing deep learning models involves collecting training data from various devices and training the models on a specific server.
However, with growing concerns about the security of user data privacy, centralized training methods are not suitable for certain cross-organizational collaborative training tasks.
Federated Learning (FL) \cite{mcmahan2017communication} has emerged as a distributed machine learning framework, allowing collaborative training of models across multiple clients while safeguarding user data privacy.
In FL, data is stored locally on each client and cannot be directly transmitted. Instead, the global model serves as the initialization model of each client, flowing among these clients. Updates from clients are then aggregated to obtain the global model for the next round on the server side \cite{abad2020hierarchical}.
FL aims to optimize a global model that can be regarded as an  ``average'' across all clients.
However, during FL training, one major practical challenge is data heterogeneity. Local data generated or collected by different clients may come from various sources, which results in diverse distribution between clients. In this case, training a single global model may struggle to efficiently adapt to the diverse data distributions across all clients.

In the literature, many works have delved into attaining ``personalized models'' for each client to alleviate potential client drift caused by heterogeneous data and most of them can achieve excellent personalized performance for each client \cite{kulkarni2020survey,wu2020fedhome,shen2022cd2,fallah2020personalized,yu2023communication,zhao2018federated}.
However, these methods typically assume a balanced global data distribution. In real-world applications, collected data often exhibits a long-tailed distribution, where most classes have very few samples (referred to as tailed classes), while a few classes contain the majority of samples (referred to as head classes) \cite{jin2023long}. It is reasonable to take into account the presence of a long-tailed distribution in PFL.
In a PFL setting characterized by both data heterogeneity and a globally long-tail distribution, the majority of clients often are in extremely imbalanced states. This imbalance leads to a significant decline in the overall performance of personalized models.
Specifically, due to data heterogeneity, there may be significant differences in the imbalance distribution among different clients. The imbalance rates or minority classes may vary across clients and diverge from the global distribution.
In this case, the global model is underrepresented in global-tailed classes due to the scarcity of samples and can only contribute limited knowledge to PFL models in identifying global-tailed classes. In addition, PFL models may also overfit some classes with a small number of samples of each client, resulting in poor PFL model performance. 
So, it becomes exceptionally challenging to learn personalized models in the PFL setting with long-tailed data, especially for the tailed classes.
FedAFA \cite{lu2023personalized} proposes to utilize the data from the majority classes of each client to generate adversarial features for its minority classes.
Although it provides a solution for PFL on the joint problem, its performance is severely affected by the network architecture and layers employed for adversarial augmentation, resulting in weak robustness.
Furthermore, due to privacy considerations, the global long-tailed distribution is unknown. 
Many long-tailed learning methods \cite{cui2019class,cao2019learning} cannot be directly employed to tackle this issue.

In this paper, we propose an Expert Collaborative Learning (ECL) framework to address the joint problem of data heterogeneity and long-tailed distribution in PFL.
Importantly, our method does not necessitate additional information beyond the standard FL setup, such as the label distribution for each client, ensuring privacy security to a large degree.
Specifically, we allocate multiple expert models to each client, and each expert is assigned a distinct training subset. Compared to the degree of imbalance in the corresponding client, the imbalance within each expert is relatively smaller,  which enables each class to receive sufficient attention and training. Especially, minority classes can also become the dominant class for specific experts and be well-trained. 
Each expert specializes in predicting specific categories. By combining the predictive strengths of each expert, we form the final prediction outputs.
Therefore, our approach is simple yet effective and demonstrates robustness to network architecture.
We conduct experiments on various heterogeneous and long-tailed benchmarks, and the results show that ECL outperforms the state-of-the-art PFL and long-tailed methods.
\section{Related work}

\subsection{Personalized Federated Learning}
Due to the non-IID data distribution among clients, the single global model cannot adapt to the distribution of each client \cite{wang2019federated}.
PFL can provide a customized model for each client. A series of works have been proposed about PFL, which are mainly two strategies \cite{tan2022towards}.
The first strategy is \textit{Global Model Personalization}, whose aim is to personalize the trained global FL model through a local adaptation step. 
Some methods, such as \cite{jeong2018communication,wu2020fedhome}, utilize data augmentation techniques to generate additional data for each client to produce an IID dataset. Another line of research focuses on the design of FL client selection mechanisms for sampling from a more homogeneous distribution \cite{wang2020optimizing,chai2020tifl}.  
Furthermore, Ditto \cite{li2021ditto} adds a regularization term on the local optimization objective, which encourages the personalized models to be close to the optimal global model.
FedBN \cite{li2021fedbn} proposes to keep the Batch Normalization layers on the client and exclude it from federated aggregation.

The second strategy is \textit{Learning Personalized Models}, aiming to modify the FL model aggregation process. 
The parameter decoupling method is an intuitive and effective approach in PFL. It entails partitioning the overall model parameters into global and personalized components. Personalized parameters are trained locally on the clients and are not shared with the FL server.
FedPer \cite{arivazhagan2019federated} retains the classifier parameters as personalized parameters and the parameters of the feature extractor as globally shared parameters.
Instead, LG-FedAvg \cite{liang2020think} shares the global classifier parameters and retains the feature extractor as personalized parameters. 
PFLEGO \cite{nikoloutsopoulos2022personalized} is based on the framework of FedPer and improves the optimization process. 
CD2-pFed \cite{shen2022cd2} uses channel decoupling to divide model parameters into global and personalized ones by channel. 
pFedLA \cite{ma2022layer} leverages hypernetworks to train layer aggregation parameters for each client to acquire personalized models.
Additionally, PFL methods based on knowledge distillation, such as \cite{he2020group,bistritz2020distributed}, achieve promising performance by acquiring a stronger global model or directly assisting in obtaining improved personalized models.
However, these methods assume a balanced global data distribution. When the global distribution follows a long-tailed pattern, the insufficient samples for tail classes result in inadequate learning for these classes, leading to a sharp decline in the performance of personalized models.
Recent works \cite{shang2022federated,zeng2023global} have been proposed to deal with Federated long-tailed problems. 
However, the personalized federated long-tailed problem is still an area that requires further exploration.
Though FedAFA \cite{lu2023personalized} has been proposed to solve the problem of data heterogeneity and long-tailed distribution in PFL through feature augmentation, it is severely affected by the network architecture and layers used for adversarial augmentation. In this paper, we explore this issue and provide a simple yet effective algorithm for tackling it through expert collaboration.
\subsection{Long-Tailed Visual Recognition}
Long-tailed visual recognition aims to improve the accuracy of tailed classes while minimizing the damage to the accuracy of head classes. \textit{Re-balancing} is the most common practice in long-tailed visual recognition, which aims to re-balance the distribution by over-sampling tailed classes, under-sampling head classes, or re-weighting the cost function according to the frequency or importance of samples \cite{japkowicz2002class,cui2019class,cao2019learning,zhou2020bbn}. Specifically, BBN \cite{zhou2020bbn} finds that the re-balancing method would damage the classifier learning to a certain extent and provides a bilateral branch network to remedy it. 
Other methods include \textit{adopting logit adjustment} by post-processing the model prediction according to class frequencies \cite{hong2021disentangling,li2022key}.
cRT \cite{kang2019decoupling} improved the model's performance by decoupling the model into feature learning and classifier learning to obtain better representation learning. Moreover, 
\textit{data augmentations} can help re-balance the distribution by transferring head classes information to tailed classes \cite{kim2020m2m}, generating tailed classes' features by estimating the distribution of tailed classes \cite{vigneswaran2021feature}, or using prior knowledge \cite{zang2021fasa}. The current methods based on augmentation can be divided into sample level and feature level. SHIKE \cite{jin2023long} fuses the features of the shallow parts and the deep part in each expert to improve representation in the first stage and utilizes the balanced softmax cross-entropy \cite{ren2020balanced} to retrain the classifier in the second stage.

Recently, methods based on a mixture of experts (MoE) have shown the potential to improve performance by leveraging the knowledge of multiple models. LFME \cite{xiang2020learning} divides long-tailed data into different groups for different experts to reduce the imbalance. 
NCL \cite{li2022nested} combines self-supervised contrastive and knowledge distillation to train multiple complete networks to handle long-tailed distributions. Nevertheless, most of them are proposed for centralized training and require global class distribution. During FL training, it is infeasible to gather the class distribution of each client to obtain the global class distribution, which makes the vast majority of long-tail learning methods not applicable to FL scenarios.

\section{Proposed Method}
\subsection{Problem Setting}
In this paper, we adopt a typical PFL system with $K$ clients participating in the training process by holding their respective local datasets $\mathcal{D}$ = $\left\{\mathcal{D}_1, \mathcal{D}_2, \cdots \mathcal{D}_K \right \}$ with $\mathcal{D}_k$ =  $\left\{\left ( \boldsymbol{x}_i,y_i  \right ) \right \}_{i=1}^{n_k}$ for client $k$, where $ \boldsymbol{x}_i$ represents the $i$-th sample, $y_i \in \left\{1, 2, \cdots C \right \}$ is the corresponding label, and $n_k$ is the number of samples. Let $n^c$ be the number of samples of class $c$, we have $n^c=\Sigma_{i=1}^{K} {n_k^c}$, and ${n_k^c}$ is the number of samples of class $c$ in client $k$.
In this paper, we consider $\mathcal{D}$ as a long-tailed distribution. Without loss of generality, we follow the common long-tailed distribution setting where the number of training data decreases as the class index increases, i.e., $n^1 > n^2 > \cdots > n^k$. 
Let $\phi_{\boldsymbol{w}}$ denote the global model with parameters $ \boldsymbol{w}$, which consists of two components: (1) a feature extractor $f$, 
where we use ResNet\cite{he2016deep} as an example, $f$ has several learnable blocks and map each example $\boldsymbol{x}$ to a $d$-dim vector. (2) a classifier $h$, which is composed of a fully connected layer and outputs predictive logits of feature vectors.
FedAvg \cite{mcmahan2017communication} is the most used FL algorithm, and it is the foundation of most PFL methods. In round $t$, clients participating in federated training download $\boldsymbol{w}$ from the server and update $\boldsymbol{w}$ using local data. Take client $k$ as an example:
 \begin{equation}
\boldsymbol{w}_k^{t+1}= \boldsymbol{w}_k^{t}-\lambda\nabla\mathcal{L}_k\left ( \mathcal{D}_k;\boldsymbol{w}_k^{t}\right).
\end{equation}
After local updating, clients in the set of $S_t$ upload the updated model $ \boldsymbol{w}_k^t $, and the server aggregate uploaded models by average weights:
\begin{equation}
\boldsymbol{w}^{t+1}= \sum_{k\in S_t}\frac{|\mathcal{D}_k|}{\Sigma_{i\in S_t}|\mathcal{D}_i|}\boldsymbol{w}_k^{t+1},
\end{equation}
where $S_t$ represents the clients participating in the training in round $t$.

\subsection{The ECL Framework}\label{AA}
Given the limited data on each client, which is insufficient for training a high-performance model individually, personalized federated learning is based on federated training. It aims to utilize the knowledge of other clients to help train personalized models while protecting the privacy of each client's data.
ECL is a simple yet effective method based on FedAvg to help train personalized expert models by fully utilizing the backbone network of the global model as a carrier of "knowledge" for other clients.
It is based on the following intuitions.
Firstly, the global model is trained indirectly on all local datasets, which results in better generalization and robustness than training on each client's local data alone. Second, the classifier is more negatively affected by long-tailed distributions, while the impact on the feature extractor's representational ability is less \cite{zhou2020bbn,kang2019decoupling}. This observation holds in federated learning of heterogeneous and long-tailed distributions \cite{shang2022federated}. 
Leveraging the backbone network of the global model obtained through the FedAvg method is both reasonable and feasible in the training process of personalized models. Additionally, our approach is orthogonal to the training of the global model, where FedAvg can also be substituted with other methods capable of obtaining superior feature representations.

\begin{figure}[!t]
    \centering
    \includegraphics[scale=.41]{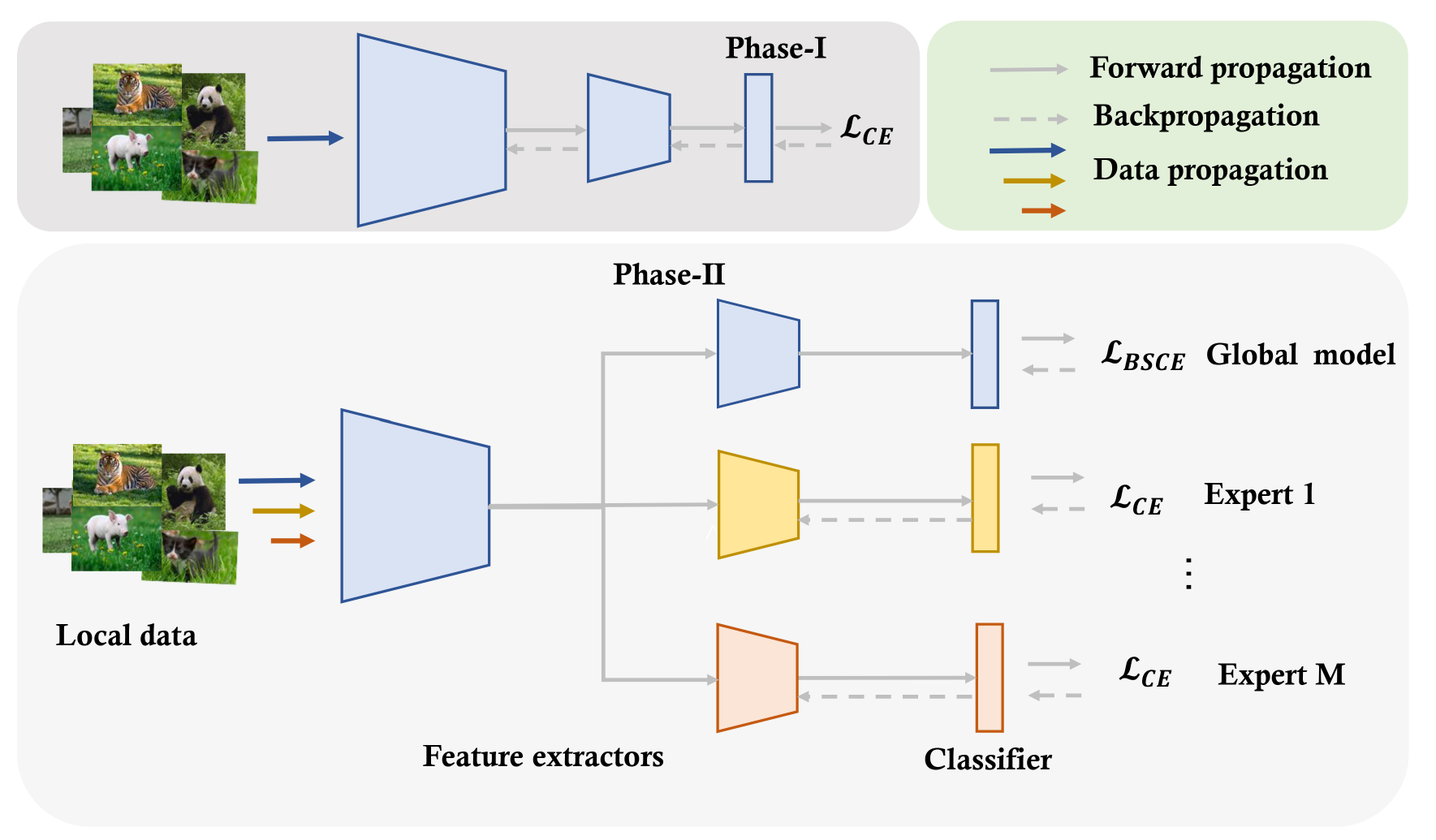}
    \caption{The framework of proposed ECL on each client. }
    \label{framework}
\end{figure}
The framework of ECL on client-sides is illustrated in Fig.~\ref{framework}. ECL consists of two stages. In Phase \uppercase\expandafter{\romannumeral1}, we leverage FedAvg to obtain a converged global model. In Phase \uppercase\expandafter{\romannumeral2}, we aim to acquire personalized expert models.
Specifically, we assign multiple expert models for each client, with these expert models sharing the same architecture as the global model, only reusing the backbone network of the global model. We sort each client's data based on sample count in descending order and sequentially divide it into different groups, each serving as the training dataset for the corresponding expert.
In this context, each expert exhibits a notably smaller degree of imbalance when compared to the imbalance present within the corresponding client, which ensures adequate training for every class.
Particularly, minority classes also have the potential to become the dominant class for a specific expert, receiving more attention and effective training.
The global model exhibits good generalization and robustness, but its classifier is more attuned to the global data distribution. Hence, we retrain the classifier of the global model. During the inference process, we combine the predicted logits from the retrained classifier of the global model with the predicted logits from multiple expert models to obtain the final prediction output.
Next, three key steps in ECL, namely, Data Grouping, Personalized Multi-Expert Model Optimization, and Multi-expert Output Aggregation in the inference stage, will be elucidated in detail.

\begin{algorithm}[!t]
\caption{: Training process of ECL}   
\label{pecl}
\textbf{Input:} Communication round $T$; Local epochs $R$; Initial global model $\mathbf{w}^0$; Personalized expert model collection $\left\{ \mathbf{v}_k^1,\mathbf{ v}_k^2,\cdots,\mathbf{v}_k^M \right\}$, where, $k \in \left[ K \right]$, $M$ represents the number of expert models, $K $ represents the collection of clients

\textbf{Return:} $\left\{ \mathbf{w}^{T}_k,\mathbf{v}_k^1,\mathbf{v}_k^2,\cdots,\mathbf{v}_k^M \right\}$  $k \in \left[ K \right]$
\begin{algorithmic}[1] 
\STATE \textbf{Phase I:}

\FOR{$t = 0$ $to$ $T - 1$}
\STATE Randomly select a group of clients $S_t$ and distribute $\mathbf{w}^t$ to the selected clients;
\FOR{ Client $k$ $\in$ $ S_t $ in parallel }
\STATE $\mathbf{w}^t_{k,0}$ $\gets$ $\mathbf{w}^t$;
\FOR{$r=0$ $ to$ $ R - 1$}
\STATE  Update $\mathbf{w}^t_{k,r}$ by using Eq. (1);
\ENDFOR
\STATE Upload $\mathbf{w}^{t+1}_{k}$ to the server;
\ENDFOR
\STATE Obtain $\mathbf{w}^{t+1}$ on the server by using Eq. (2);
\ENDFOR
\STATE  \textbf{Phase II:}
\FOR{ Client $k$ $\in$ $ K $ in parallel }
\STATE {Group the client’s data according to the data grouping method in Section 3.2 $|\mathcal{C}_1 \cup \mathcal{C}_2 \cup \ldots \cup \mathcal{C}_M|$};
\STATE Use the global model $\mathbf{w}^{T}$ obtained in Phase I to initialize the expert models;
\STATE {Update the global model$\mathbf{w}^{T}_k$ by using Eq. (3)};
\FOR{ Each expert model $m$   in parallel}
\STATE {Update $\mathbf{v}_k^m$ by using Eq. (4) and the corresponding dataset $\mathcal{C}_m$};
\ENDFOR
\ENDFOR
\end{algorithmic}
\end{algorithm}
\noindent{\bf Data grouping.} For the $k$-th client, assuming its dataset comprises a total of $C$ classes, and there are $M$ expert models involved in training.
We sort the dataset $\mathcal{D}_k$ in descending order based on the sample quantity of each class.
Each expert is assigned a training subset $ \mathcal{C}_i$,
where $ |\mathcal{C}_1 \cup \mathcal{C}_2 \cup \ldots \cup \mathcal{C}_M| = C $ and $\mathcal{C}_i \cap \mathcal{C}_j=\emptyset$.
Without loss of generality, we assume that $C$ is a multiple of $K$, and each $\mathcal{C}_i$ corresponds to a continuous range of $ \frac{C}{K}$ classes.

\noindent{\bf Personalized Multi-Expert Model Optimization.}
In Phase \uppercase\expandafter{\romannumeral2}, we train and learn personalized expert models for each client based on the global model obtained in Phase \uppercase\expandafter{\romannumeral1}.
Specifically, since the backbone network of the global model encapsulates rich 'knowledge' from other clients, we use the backbone network parameters of the global model to initialize the backbone network of the expert models.
For the global model, assuming it is on client $k$, we retrain the classifier using the entire dataset $D_k$ for a balanced classifier tailored to the local distribution of client $k$. Therefore, we freeze the backbone network parameters and utilize the Balanced Softmax Cross Entropy (BSCE) \cite{ren2020balanced} as the loss function $\mathcal L_{bsce}$ to optimize the classifier:
\begin{equation}
\mathcal L_{bsce} = -\sum_{m=1}^M\log\left( \frac{n_k^y\exp\left( z^m\right)}{\Sigma_{j=1}^Cn_k^j\exp\left( z_j^m\right)}\right).
\end{equation}
For the expert models, we assign different subsets for each expert. Taking ResNet as an example, we typically update only the last residual block and the classifier layer for the majority of expert models. However, for expert models with tailed class data for the client,  usually the last expert ($M$-th expert) we update only the classifier parameters using a class-balanced local dataset $\mathcal{C}_M$.
We update the expert models using the following cross-entropy loss function:
\begin{equation}
\mathcal L_{ce} = -\sum_{m=1}^M\log\left( \frac{\exp\left( z^m\right)}{\Sigma_{j=1}^C\exp\left( z_j^m\right)}\right).
\end{equation}

\noindent{\bf{ Multi-expert Output Aggregation.}} In the reference stage, the output logits of expert models are $\boldsymbol{z}$, which is adjusted to $\bar {\boldsymbol{z}}$.
\begin{equation}
\bar {\boldsymbol{z}} =\frac{\left \| \boldsymbol{u} \right \|^2}{\left \|\boldsymbol{u}_0 \right \|^2} \cdot \boldsymbol{z},
\end{equation}
where $\left \| \boldsymbol{u} \right \|$ and $\left \|\boldsymbol{u}_0 \right \|$ are the L2 norm of the classifier weights of the expert model and global model, respectively.
Finally, the corresponding expert and the global model jointly obtain the prediction logits of class $c$ by
\begin{equation}
\boldsymbol{o}^c =\lambda\bar {\boldsymbol{z}}^c+ \left(1 - \lambda\right) \boldsymbol{z}_0^c,
\end{equation}
where $\boldsymbol{o}^c$ is the aggregated logits, $\boldsymbol{z}_0^c$ is the output logit of class $c$ by the global model. 
$\lambda$ is the scaling factor. When $\lambda = 0$, the final logit of class $c$ is the prediction output of the global model. When $\lambda = 1$, it is the predicted output of the corresponding expert model.
 Our algorithm is summarized in Algorithm 1.

\begin{table}[!h]
\caption{Top-1 test accuracy (\%) for ECL and 
compared PFL methods on CIFAR10/100-LT with different IFs. (·) indicates the comparison to the Local, where \textcolor{green!80!black}{increase} and \textcolor{red}{decrease} are represented by green and red,respectively. The results of ECL are presented in bold.}
\label{tab1}
\renewcommand\arraystretch{1}
\begin{center}
\setlength{\tabcolsep}{0.1mm}
\renewcommand{\arraystretch}{1.2}
\begin{tabular}{l|>{\centering\arraybackslash}p{1.5cm}> 
    {\centering\arraybackslash}p{1.5cm}>{\centering\arraybackslash}p{1.5cm}|>{\centering\arraybackslash}p{1.5cm}>{\centering\arraybackslash}p{1.5cm}>{\centering\arraybackslash}p{1.5cm}}
\toprule

Dataset & \multicolumn{3}{c|}{CIFAR10-LT} & 
\multicolumn{3}{c}{CIFAR100-LT}\\
\midrule
\centering
 IF&\multicolumn{1}{c}{10}&\multicolumn{1}{c}{50}&\multicolumn{1}{c|}{100}&\multicolumn{1}{c}{10}&\multicolumn{1}{c}{50}&\multicolumn{1}{c}{100}\\ 
\midrule
Local & 52.88 &49.06 &47.86& 21.93&22.15 &22.56  \\
FedAvg-FT \cite{wang2019federated}  &53.75 &50.33 & 50.47&25.09 &27.17 & 26.02 \\
 & \textcolor{green!80!black}{(+0.87)} & \textcolor{green!80!black}{(+1.27)} & \textcolor{green!80!black}{(+2.61)} & \textcolor{green!80!black}{(+3.16)} & \textcolor{green!80!black}{(+5.02)} & \textcolor{green!80!black}{(+3.46)} \\
LG-FedAvg \cite{liang2020think} &53.44 &49.55 & 47.99&21.26 &22.70 & 22.15 \\
 & \textcolor{green!80!black}{(+0.56)} & \textcolor{green!80!black}{(+0.49)} & \textcolor{green!80!black}{(+0.13)} & \textcolor{red}{(-0.67)} & \textcolor{green!80!black}{(+0.55)} & \textcolor{red}{(-0.41)} \\
FedPer \cite{arivazhagan2019federated}  &69.51 \textcolor{green!80!black}{(+16.63)} &64.13 \textcolor{green!80!black}{(+15.07)}&62.44 \textcolor{green!80!black}{(+14.58)}&34.07 \textcolor{green!80!black}{(+12.14)}&34.31 \textcolor{green!80!black}{(+12.16)}&34.28 \textcolor{green!80!black}{(+11.72)}\\
Ditto \cite{li2021ditto} &75.37 \textcolor{green!80!black}{(+22.49)}  &66.30 \textcolor{green!80!black}{(+17.24)}&62.93 \textcolor{green!80!black}{(+15.07)}&52.85 \textcolor{green!80!black}{(+30.92)}&49.32 \textcolor{green!80!black}{(+27.17)}&48.04 \textcolor{green!80!black}{(+25.48)}\\
FedLC \cite{zhang2022federated}  & 71.06 \textcolor{green!80!black}{(+18.18)} &59.85 \textcolor{green!80!black}{(+10.79)}&57.66 \textcolor{green!80!black}{(+9.8)}&48.51 \textcolor{green!80!black}{(+26.58)}&43.39 \textcolor{green!80!black}{(+21.24)}&42.31 \textcolor{green!80!black}{(+19.75)}\\
FED-RoD \cite{chen2021bridging}  &58.49 \textcolor{green!80!black}{(+5.61)} &54.20 \textcolor{green!80!black}{(+5.14)}&51.70 \textcolor{green!80!black}{(+3.84)}&32.70 \textcolor{green!80!black}{(+10.77)}&30.65 \textcolor{green!80!black}{(+8.5)}&29.27 \textcolor{green!80!black}{(+6.71)}\\
FedBN \cite{li2021fedbn}  & 68.90 \textcolor{green!80!black}{(+16.02)}&59.77 \textcolor{green!80!black}{(+10.71)}&59.54 \textcolor{green!80!black}{(+11.68)}&43.60 \textcolor{green!80!black}{(+21.27)}&41.34 \textcolor{green!80!black}{(+19.19)}&39.73 \textcolor{green!80!black}{(+17.17)}\\
pFedLA \cite{ma2022layer} & 61.24 \textcolor{green!80!black}{(+8.36)} & 61.04 \textcolor{green!80!black}{(+11.98)}&63.08 \textcolor{green!80!black}{(+15.22)}&47.96 \textcolor{green!80!black}{(+26.03)}&47.87 \textcolor{green!80!black}{(+25.72)}&48.72 \textcolor{green!80!black}{(+26.16)}\\
\midrule
FedAvg+LWS \cite{kang2019decoupling} &74.58 \textcolor{green!80!black}{(+21.70)} &62.88 \textcolor{green!80!black}{(+13.82)}&59.00 \textcolor{green!80!black}{(+11.14)}&53.34 \textcolor{green!80!black}{(+31.41)}&49.92 \textcolor{green!80!black}{(+27.77)}&45.83 \textcolor{green!80!black}{(+23.27)}\\
FedAvg+$\tau$-norm \cite{kang2019decoupling} &79.60 \textcolor{green!80!black}{(+26.72)} &69.41 \textcolor{green!80!black}{(+20.35)}&63.52 \textcolor{green!80!black}{(+15.66)}&55.73 \textcolor{green!80!black}{(+33.80)}&51.28 \textcolor{green!80!black}{(+29.13)}&47.47 \textcolor{green!80!black}{(+24.91)}\\
FedAFA \cite{lu2023personalized} &67.08 \textcolor{green!80!black}{(+14.20)}&62.55 \textcolor{green!80!black}{(+13.49)}&59.70 \textcolor{green!80!black}{(+11.84)}&37.71 \textcolor{green!80!black}{(+15.78)}&38.82 \textcolor{green!80!black}{(+16.67)}&36.83 \textcolor{green!80!black}{(+14.27)}\\
\midrule
\textbf{ECL (2 experts)} &\textbf{86.32} \textcolor{green!80!black}{(+33.44)}&\textbf{79.45} \textcolor{green!80!black}{(+30.39)}&\textbf{76.26} \textcolor{green!80!black}{(+28.40)} &\textbf{62.46} \textcolor{green!80!black}{(+40.53)}&\textbf{61.21} \textcolor{green!80!black}{(+39.06)}&\textbf{57.96} \textcolor{green!80!black}{(+35.40)}\\ 
\textbf{ECL (3 experts)} & \textbf{85.74} \textcolor{green!80!black}{(+32.86)} &\textbf{78.98} \textcolor{green!80!black}{(+29.92)} & \textbf{75.41 }\textcolor{green!80!black}{(+27.55)}& \textbf{61.49} \textcolor{green!80!black}{(+39.56)}& \textbf{58.78} \textcolor{green!80!black}{(+36.63)}&\textbf{55.98} \textcolor{green!80!black}{(+33.42)}\\ 
\bottomrule
\end{tabular}
\end{center}
\end{table}
\begin{table}[!t]
\centering
\caption{Top-1 test accuracy  (\%) for compared PFL methods and ECL on FashionMNIST-LT and tiny-ImageNet200-LT with different IFs and $\rm \alpha$. The last column ($ \rm IF=10*$) is the result on tiny-ImageNet200-LT.}\label{tab2}
\setlength{\tabcolsep}{1.5mm}
\renewcommand{\arraystretch}{1.2}
\begin{tabular}{lccccc}
\toprule
\multirow{2}{*}{ Method}& $\rm IF=10$  & \multicolumn{3}{c}{$\rm IF=100$ } & $\rm IF=10*$ \\
\cmidrule(lr){2-2}\cmidrule(lr){3-5}\cmidrule(lr){6-6}
\multicolumn{1}{c}{}                        & \multicolumn{1}{c}{$ \rm \alpha=0.2$} & \multicolumn{1}{c}{$ \rm \alpha =0.2$} & \multicolumn{1}{c}{$ \rm \alpha=0.8$} & \multicolumn{1}{c}{$\rm \alpha=1.0$} & \multicolumn{1}{c}{$\rm \alpha=0.2$} \\
\hline
Local & 75.15  & 73.19 &80.00&80.36& 15.36\\
FedAvg-FT&81.28   &81.68 & 84.62&85.61&19.21 \\
FedPer &  82.42 & 81.07& 84.10& 85.27 &22.40\\
FedAvg+LWS& 87.58  & 82.91&87.43&86.63&28.09 \\
FedAvg+$\tau$-norm&89.12   &84.98 &88.29& 87.83& 20.40\\
\midrule
ECL (2 experts) &\textbf{91.49}   &\textbf{89.17} &\textbf{89.70} & \textbf{90.19}&\textbf{35.84}\\
\hline
\end{tabular}
\end{table}
\section{Experiments}
\subsection{Experiment Setup}
We evaluate ECL using the image classification dataset. \textbf{CIFAR10/100-LT} are subsets of the original CIFAR10/100 dataset \cite{krizhevsky2009learning} with long-tailed distribution. 
\textbf{FashionMNIST-LT} and \textbf{tiny-ImageNet200-LT} are also subsets of FashionMNIST \cite{xiao2017fashion} and tiny-ImageNet200 \cite{le2015tiny} with long-tailed distribution, respectively. 
The dataset tiny-ImageNet200 contains $200$ classes with $500$ images for each class in the training set, $50$ images for each class in the validation set, and $10,000$ images in the testing set without labels. 
The Imbalance Factor (IF) controls the degree of data imbalance in the dataset. 
In this paper, we shape long-tailed distribution with IF = 10, 50, and 100 for CIFAR10/100-LT, IF = 10 for tiny-ImageNet200-LT and IF=10 and 100 for FashionMNIST-LT. 

We employ ResNet-32 \cite{zhou2020bbn} as the backbone network for  tiny-ImageNet200-LT and CIFAR datasets, a simple CNN with two convolutional layers and three fully connected layers as the backbone network For FashionMNIST-LT. We use the Dirichlet distribution with the hyperparameter $\alpha$ to simulate the degree of data heterogeneity, as done in \cite{lin2020ensemble}. 
We set $\alpha = 0.2$ without special instructions. We adopt an SGD optimizer with a learning rate of $0.1$, a momentum of $0.9$, and a weight decay of $5\times10^{-4}$ for the first $200$ rounds. In the 200-th round, we change the learning rate to $0.01$. The total number of global communication rounds is $500$. We set the total number of clients at 20, with only 10 clients randomly selected for CIFAR datasets and FashionMNIST-LT and 8 clients for tiny-ImageNet200-LT in each round. Unless specified otherwise, the hyperparameters $\lambda=1/2$.
\subsection{Comparison with the State-of-the-art Methods}
\noindent {\bf Results on CIFAR10/100-LT.}
The comparison results are summarized in Table \ref{tab1}.
It can be seen that ECL with both two and three expert models achieve similar superior performance, outperforming existing state-of-the-art PFL methods. Among them, using two expert models yielded the best results, primarily because,  with highly heterogeneous data and long-tail distributions where each client has fewer categories, employing two expert models adequately ensures comprehensive training for each class within the local client.
Local training unexpectedly achieves almost the worst results due to a lack of generalization performance. Other PFL methods may not yield the best results as they only consider the impact of data heterogeneity without considering the influence of long-tailed distribution. Although long-tailed methods applied to the framework of PFL achieve good performance, they still have a certain gap with ECL.  FedAFA addresses the problems of long-tailed data in PFL through feature augmentation. However, the performance of FedAFA is not robust and is not as good as ECL. 

\begin{figure}[!t]
	\centering
	\begin{minipage}[b]{\linewidth}
		\centering
            \subfigure[]{\includegraphics[width=0.47\linewidth]{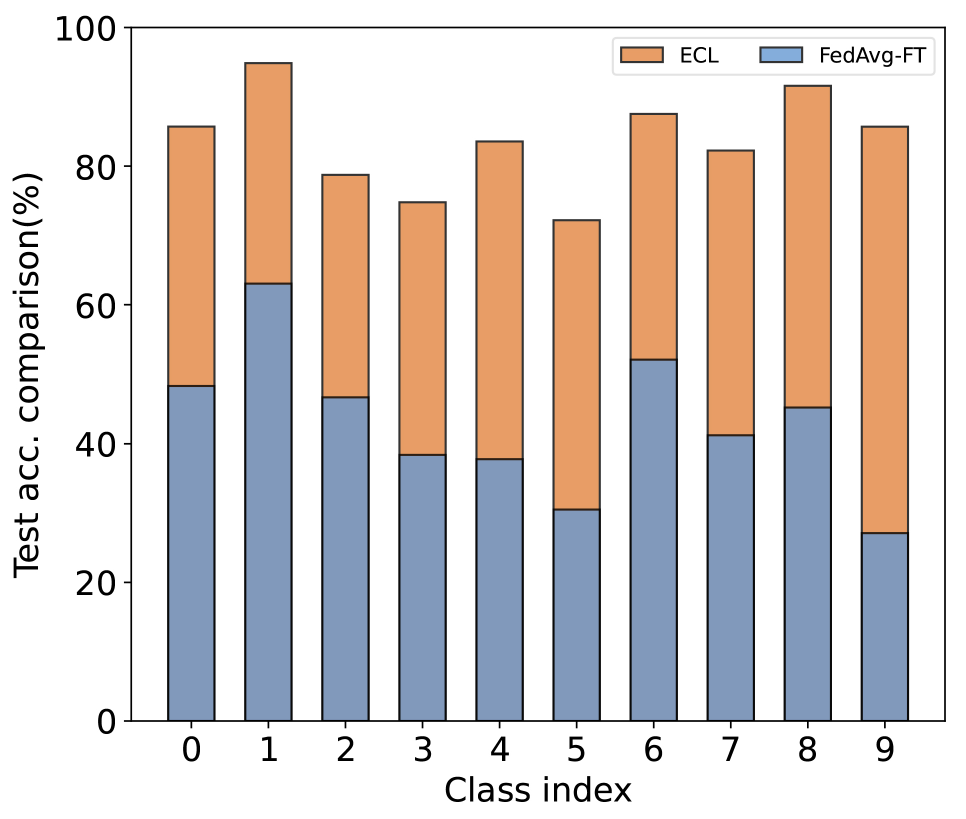}}
            \subfigure[]{
             \includegraphics[width=0.47\linewidth]{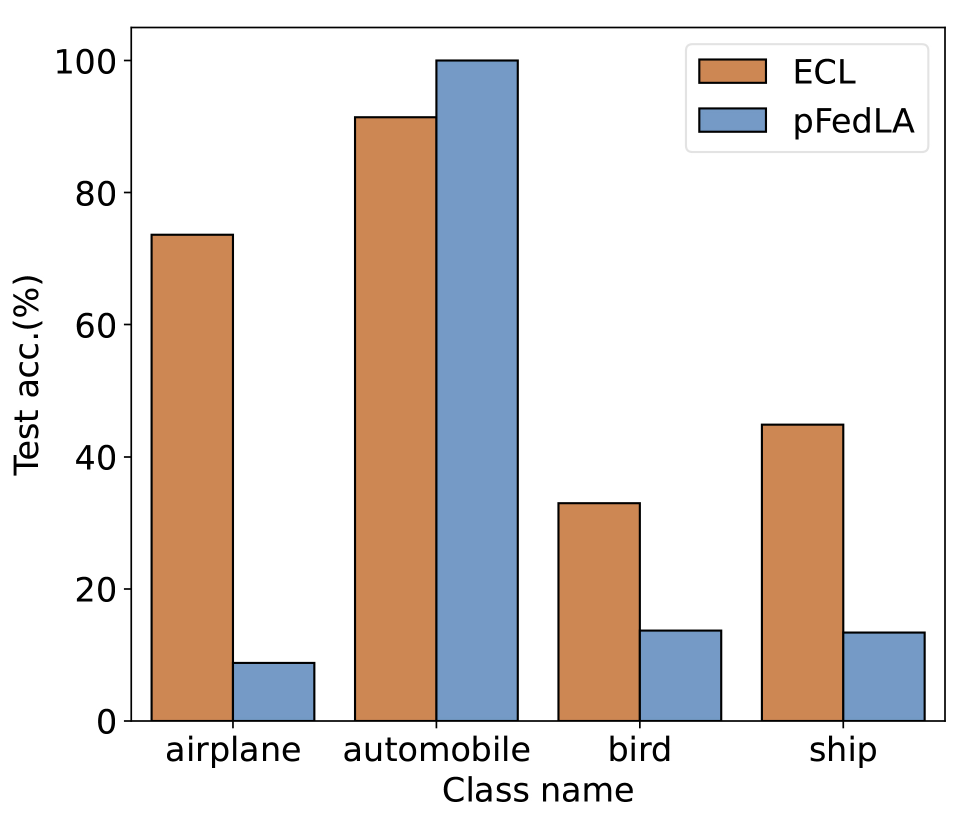}
            }
	\end{minipage}
\caption{(a) Visualization of test accuracy gap on CIFAR10-LT with IF=10. (b) Visualizing the gap in test accuracy on a client, where the numbers of samples of each class from left to right are \{1, 823, 1, 1\}}
\label{mu}
\end{figure}

To demonstrate the performance enhancement of personalized models by ECL, we evaluate its improvement compared to FedAvg-FT in each class. The results depicted in Fig.\ref{mu} (a), show that ECL exhibits significant improvements across all classes, particularly for tailed classes. We randomly select a client and compare our proposed method with pFedLA on CIFAR10-LT with IF=100. As shown in Fig.\ref{mu} (b), it shows that ECL significantly improves the performance of minority classes with a slight tolerable decrease in performance for majority classes within the client.

\noindent {\bf Results on tiny-ImageNet200-LT and FashionMNIST-LT.}
We further evaluate ECL on tiny-ImageNet200-LT and FashionMNIST-LT, with the results presented in  Table \ref{tab2}. 
In comparison to other methods,  ECL consistently achieves the best performance, underscoring its robust generalization capabilities across diverse tasks.

\subsection{Model Validation}
\noindent {\bf Effectiveness of data heterogeneity.}
We further assess the performance of ECL on FashionMNIST-LT across varying degrees of data heterogeneity. As depicted in Table \ref{tab2}, ECL consistently outperforms other methods across different values of $ \rm \alpha$. Notably, ECL exhibits even more significant performance gains when the data distribution is highly heterogeneous (with a smaller $ \rm \alpha$).

\noindent {\bf Effectiveness of output aggregation.}
\begin{table}[!t]
\caption{Comparisons of output aggregation schemes on  CIFAR10-LT}\label{tab3}
\centering
\setlength{\tabcolsep}{4mm}
\renewcommand{\arraystretch}{1.2}
\begin{tabular}{lccc}
\hline
Method & $\rm IF=10$ & $\rm IF=50$ & $\rm IF=100$  \\
\hline
with LWS (1) & 82.17 & 73.67 & 70.24 \\
without scaling (2) & 81.47 & 73.65 & 71.58 \\
with scaling (ECL) & \textbf{86.32} & \textbf{79.45} & \textbf{76.26} \\
\hline
\end{tabular}
\end{table}
We also compare various aggregation methods of output logits, and the results are reported in Table \ref{tab3}. Method (2) does not involve scaling, while method (1) scales each class by the classifier weight of corresponding expert models and the global models. Overall, merging expert models requires trade-offs, and ECL consistently achieves the best results.

\noindent{\bf Effectiveness of other existing FL optimizers.}
 Table \ref{tab4} presents the performance of ECL with FedProx, SCAFFOLD, and FedLC optimizer on CIFAR10-LT with different IFs. We observe that incorporating ECL can substantially enhance the performance of any federated learning optimizer, and using the simple and robust FedAvg method in the first stage achieves better results. 


\begin{table}[!t]
\caption{The impact of FL optimizer with ECL on CIFAR10-LT. The values in parentheses are the original test accuracy without ECL (\%).}\label{tab4}
\centering
\setlength{\tabcolsep}{2mm}
\renewcommand{\arraystretch}{1.5}
\begin{tabular}{llll}
\hline
           \multicolumn{1}{l}{Method}       & \multicolumn{1}{c}{$\rm IF=10$} & \multicolumn{1}{c}{$\rm IF=50$} & \multicolumn{1}{c}{$\rm IF=100$} \\  \hline
FedProx+ECL       & 86.28 (80.27) & 77.92 (62.09)  & 73.93 (58.30)  \\ 
SCAFFOLD+ECL      & 76.53 (63.60)     & 69.32 (58.17)  & 64.35 (50.69)  \\ 
FedLC+ECL         & 74.78 (71.06)    &   65.09 (59.85)  &  60.92 (57.66) \\ \hline
\end{tabular}
\end{table}

\section{Conclusion}
In this paper, we have proposed an Expert Collaboration Learning (ECL) framework to address the joint problems of heterogeneous and long-tailed data in PFL. 
In highly heterogeneous long-tail data, there is a significant imbalance within each client, where minority class data tends to be overlooked, leading to an overall decline in personalized performance. We allocate multiple experts to each client, with each expert corresponding to a different training subset. Compared to the imbalance within the respective clients, this strategy significantly reduces the imbalance within each expert, ensuring that each class receives sufficient training and attention. Ultimately, we perform a weighted summation of the predicted logits from the global model and expert models to obtain the final prediction output.
Extensive experimental results show that ECL outperforms the state-of-the-art PFL methods under different settings.

%
%
%

\begin{thebibliography}{8}
\bibitem{zhao2023federated}
Zhao, Z., Yang, F., Liang, G.: Federated Learning Based on Diffusion Model to Cope with Non-IID Data. In: Chinese Conference on Pattern Recognition and Computer Vision. pp. 220-231. (2023)
\bibitem{mcmahan2017communication}
McMahan, B., Moore, E., Ramage, D., Hampson, S., y Arcas, B. A.: Communication-efficient learning of deep networks from decentralized data. . In:
Artificial Intelligence and Statistics, pp. 1273–1282. PMLR (2017)
\bibitem{abad2020hierarchical}
Abad, M. S. H., Ozfatura, E., Gunduz, D., Ercetin, O.: Hierarchical federated learning across heterogeneous cellular networks. In: IEEE International Conference on Acoustics, Speech and Signal Processing, pp. 8866-8870 (2020)
\bibitem{kulkarni2020survey}
Kulkarni, V., Kulkarni, M., Pant, A.: Survey of personalization techniques for federated learning. In: 2020 Fourth World Conference on Smart Trends in Systems, Security and Sustainability, pp. 794-797 (2020)
\bibitem{wu2020fedhome}
Wu, Q., Chen, X., Zhou, Z., Zhang, J.:  Fedhome: Cloud-edge based personalized federated learning for in-home health monitoring. IEEE Transactions on Mobile Computing, vol. 21(8), pp. 2818-2832 (2020)
\bibitem{shen2022cd2}
Shen, Y., Zhou, Y., and Yu, L.: CD2-pFed: Cyclic Distillation-guided Channel Decoupling for Model Personalization in Federated Learning. In: Proceedings of the IEEE/CVF Conference on Computer Vision and Pattern Recognition, pp. 10041–10050 (2022)
\bibitem{fallah2020personalized}
Fallah, A., Mokhtari, A., Ozdaglar, A.: Personalized federated learning: A meta-learning approach. arXiv preprint arXiv:2002.07948 (2020)
\bibitem{yu2023communication}
Yu, F., Lin, H., Wang, X., Garg, S., Kaddoum, G., Singh, S., Hassan, M. M.: Communication-efficient personalized federated meta-learning in edge networks. IEEE Transactions on Network and Service Management  (2023)
\bibitem{zhao2018federated}
Zhao, Y., Li, M., Lai, L., Suda, N., Civin, D., Chandra, V.: Federated learning with non-iid data. arXiv preprint arXiv:1806.00582 (2018)
\bibitem{jin2023long}
Jin, Y., Li, M., Lu, Y., Cheung, Y. M., Wang, H.:  Long-tailed visual recognition via self-heterogeneous integration with knowledge excavation. In: Proceedings of the IEEE/CVF Conference on Computer Vision and Pattern Recognition, pp. 23695-23704 (2023)
\bibitem{lu2023personalized}
Lu, Y., Qian, P., Huang, G., Wang, H.: Personalized Federated Learning on Long-Tailed Data via Adversarial Feature Augmentation. In: IEEE International Conference on Acoustics, Speech and Signal Processing, pp. 1-5 (2023)
\bibitem{cui2019class}
Cui, Y., Jia, M., Lin, T. Y., Song, Y., Belongie, S.: Class-balanced loss based on effective number of samples. In: Proceedings of the IEEE/CVF Conference on Computer Vision and Pattern Recognition,  pp. 9268-9277 (2019)
\bibitem{cao2019learning}
Cao, K., Wei, C., Gaidon, A., Arechiga, N., Ma, T.: Learning imbalanced datasets with label-distribution-aware margin loss. In: Advances in Neural Information Processing Systems, vol. 32 (2019)
\bibitem{wang2019federated}
Wang, K., Mathews, R., Kiddon, C., Eichner, H., Beaufays, F., Ramage, D.: Federated evaluation of on-device personalization. arXiv preprint arXiv:1910.10252 (2019)
\bibitem{tan2022towards}
Tan, A. Z., Yu, H., Cui, L., Yang, Q.:  Towards personalized federated learning. IEEE Transactions on Neural Networks and Learning Systems (2022)
\bibitem{jeong2018communication}
Jeong, E., Oh, S., Kim, H., Park, J., Bennis, M., Kim, S. L.:  Communication-efficient on-device machine learning: Federated distillation and augmentation under non-iid private data. arXiv preprint arXiv:1811.11479 (2018)
\bibitem{wang2020optimizing}
Wang, H., Kaplan, Z., Niu, D.,  Li, B.: Optimizing federated learning on non-iid data with reinforcement learning. In: Proceedings of the IEEE Conference on Computer Communications, pp. 1698-1707. IEEE  (2020)
\bibitem{chai2020tifl}
Chai, Z., Ali, A., Zawad, S., Truex, S., Anwar, A., Baracaldo, N., Cheng, Y.: Tifl: A tier-based federated learning system. In: the 29th International Symposium on High-performance Parallel and Distributed Computing, pp. 125-136 (2020)
\bibitem{li2021ditto}
Li, T., Hu, S., Beirami, A., Smith, V.: Ditto: Fair and robust federated learning through personalization. In: International Conference on Machine Learning,  pp. 6357-6368. PMLR (2021)
\bibitem{li2021fedbn}
Li, X., Jiang, M., Zhang, X., Kamp, M., Dou, Q.:Fedbn: Federated learning on non-iid features via local batch normalization. arXiv preprint arXiv:2102.07623  (2021)
\bibitem{arivazhagan2019federated}
Arivazhagan, M. G., Aggarwal, V., Singh, A. K., Choudhary, S.: Federated learning with personalization layers. arXiv preprint arXiv:1912.00818 (2019)
\bibitem{liang2020think}
Liang, P. P., Liu, T., Ziyin, L., Allen, N. B., Auerbach, R. P., Brent, D., Morency, L. P.: Think locally, act globally: Federated learning with local and global representations. arXiv preprint arXiv:2001.01523 (2020)
\bibitem{nikoloutsopoulos2022personalized}
Nikoloutsopoulos, S., Koutsopoulos, I., Titsias, M. K. Personalized Federated Learning with Exact Stochastic Gradient Descent. arXiv preprint arXiv:2202.09848 (2022)
\bibitem{ma2022layer}
Ma, X., Zhang, J., Guo, S., Xu, W.:Layer-wised model aggregation for personalized federated learning. In: Proceedings of the IEEE/CVF Conference on Computer Vision and Pattern Recognition, pp. 10092-10101 (2022)
\bibitem{he2020group}
He, C., Annavaram, M., Avestimehr, S.: Group knowledge transfer: Federated learning of large cnns at the edge. In: Advances in Neural Information Processing Systems, vol. 33, pp. 14068-14080  (2020)
\bibitem{bistritz2020distributed}
Bistritz, I., Mann, A., Bambos, N.: Distributed distillation for on-device learning. In: Advances in Neural Information Processing Systems, vol. 33, pp. 22593-22604  (2020)
\bibitem{shang2022federated}
Shang, X., Lu, Y., Huang, G., Wang, H.: Federated learning on heterogeneous and long-tailed data via classifier re-training with federated features. arXiv preprint arXiv:2204.13399 (2022)
\bibitem{zeng2023global}
Zeng, Y., Liu, L., Liu, L., Shen, L., Liu, S., Wu, B.: Global balanced experts for federated long-tailed learning. In: Proceedings of the IEEE/CVF International Conference on Computer Vision, pp. 4815-4825 (2023)
\bibitem{japkowicz2002class}
Japkowicz, N., Stephen, S.:  The class imbalance problem: A systematic study. Intelligent data analysis, 6(5), 429-449 (2002)
\bibitem{zhou2020bbn}
Zhou, B., Cui, Q., Wei, X. S., Chen, Z. M.: Bbn: Bilateral-branch network with cumulative learning for long-tailed visual recognition. In: Proceedings of the IEEE/CVF Conference on Computer Vision and Pattern Recognition, pp. 9719-9728 (2020)
\bibitem{hong2021disentangling}
Hong, Y., Han, S., Choi, K., Seo, S., Kim, B., Chang, B.: Disentangling label distribution for long-tailed visual recognition. In: Proceedings of the IEEE/CVF Conference on Computer Vision and Pattern Recognition, pp. 6626-6636 (2021)
\bibitem{li2022key}
Li, M., Cheung, Y. M., Hu, Z.:  Key point sensitive loss for long-tailed visual recognition. IEEE Transactions on Pattern Analysis and Machine Intelligence, 45(4), 4812-4825 (2022)
\bibitem{kang2019decoupling}
Kang, B., Xie, S., Rohrbach, M., Yan, Z., Gordo, A., Feng, J., Kalantidis, Y.: Decoupling representation and classifier for long-tailed recognition. arXiv preprint arXiv:1910.09217 (2019)
\bibitem{kim2020m2m}
Kim, J., Jeong, J., Shin, J.:  M2m: Imbalanced classification via major-to-minor translation. In: Proceedings of the IEEE/CVF Conference on Computer Vision and Pattern Recognition, pp. 13896-13905 (2020)
\bibitem{vigneswaran2021feature}
Vigneswaran, R., Law, M. T., Balasubramanian, V. N., Tapaswi, M.:  Feature generation for long-tail classification. In: the twelfth Indian Conference on Computer Vision, Graphics and Image Processing, pp. 1-9 (2021)
\bibitem{zang2021fasa}
Zang, Y., Huang, C., Loy, C. C.:  Fasa: Feature augmentation and sampling adaptation for long-tailed instance segmentation. In: Proceedings of the IEEE/CVF International Conference on Computer Vision, pp. 3457-3466 (2021)
\bibitem{ren2020balanced}
Ren, J., Yu, C., Ma, X., Zhao, H., Yi, S.: Balanced meta-softmax for long-tailed visual recognition. In: Advances in Neural Information Processing Systems, vol. 33, pp. 4175-4186 (2020)
\bibitem{xiang2020learning}
Xiang, L., Ding, G., Han, J.: Learning from multiple experts: Self-paced knowledge distillation for long-tailed classification. In: Proceedings of the European Conference Computer Vision,  pp. 247-263. (2020)
\bibitem{li2022nested}
Li, J., Tan, Z., Wan, J., Lei, Z., Guo, G.: Nested collaborative learning for long-tailed visual recognition. In: Proceedings of the IEEE/CVF Conference on Computer Vision and Pattern Recognition,  pp. 6949-6958 (2022)
\bibitem{he2016deep}
He, K., Zhang, X., Ren, S., Sun, J.: Deep residual learning for image recognition. In: Proceedings of the IEEE Conference on Computer Vision and Pattern Recognition, pp. 770-778 (2016)


\bibitem{krizhevsky2009learning}
Krizhevsky, A., Hinton, G.: Learning multiple layers of features from tiny images. Technical Reports (2009)
\bibitem{xiao2017fashion}
Xiao H., Rasul K., Roland Vollgraf.: Fashion-mnist: a novel image dataset for benchmarking machine learning algorithms. ArXiv, abs/1708.07747 (2017)
\bibitem{le2015tiny}
Le, Y., Yang, X.: Tiny imagenet visual recognition challenge. CS 231N, 7(7), 3 (2015)
\bibitem{lin2020ensemble}
Lin, T., Kong, L., Stich, S. U., Jaggi, M. Ensemble distillation for robust model fusion in federated learning. In:  Advances in Neural Information Processing Systems, vol. 33, pp. 2351-2363 (2020)
\bibitem{chen2021bridging}
Chen, H. Y., Chao, W. L. On bridging generic and personalized federated learning for image classification. arXiv preprint arXiv:2107.00778 (2021)









\bibitem{zhang2022federated}
Zhang, J., Li, Z., Li, B., Xu, J., Wu, S., Ding, S., Wu, C.: Federated learning with label distribution skew via logits calibration. In: International Conference on Machine Learning,  pp. 26311-26329. PMLR (2022)

\end{thebibliography}
%

\end{document}